\documentclass[12pt, onecolumn]{IEEEtran}

\usepackage[utf8]{inputenc}  
\usepackage{amsmath, amssymb} 
\usepackage{graphicx}        
\usepackage{hyperref}        
\usepackage{cite}            

\title{Analytical Results for Two Exponential Family Distributions in Hierarchical Dirichlet Processes}
\author{
\IEEEauthorblockN{Naiqi Li, Tsinghua University\\
\small linaiqi.thu@gmail.com
}
}

\begin{document}
\maketitle

\begin{abstract}
The Hierarchical Dirichlet Process (HDP)  provides a flexible Bayesian nonparametric framework for modeling grouped data with a shared yet unbounded collection of mixture components. While existing applications of the HDP predominantly focus on the Dirichlet–multinomial conjugate structure, the framework itself is considerably more general and, in principle, accommodates a broad class of conjugate prior–likelihood pairs. In particular, exponential family distributions offer a unified and analytically tractable modeling paradigm that encompasses many commonly used distributions.
In this paper, we investigate analytic results for two important members of the exponential family within the HDP framework: the Poisson distribution and the normal distribution. We derive explicit closed-form expressions for the corresponding Gamma–Poisson and Normal–Gamma–Normal conjugate pairs under the hierarchical Dirichlet process construction. Detailed derivations and proofs are provided to clarify the underlying mathematical structure and to demonstrate how conjugacy can be systematically exploited in hierarchical nonparametric models.
Our work extends the applicability of the HDP beyond the Dirichlet–multinomial setting and furnishes practical analytic results for researchers employing hierarchical Bayesian nonparametrics.
\end{abstract}

\section{Introduction}

The Dirichlet process (DP) \cite{teh2017dirichlet,li2019tutorial} provides a principled framework for mixture models with an unbounded number of components. By placing a Dirichlet process prior on the mixing distribution, DP mixture models allow the complexity of the model to grow with the data, thereby avoiding the need to pre-specify the number of clusters. This property makes DP-based clustering particularly attractive in applications where the underlying structure is unknown or potentially infinite-dimensional.

While the Dirichlet process is well suited for modeling a single group of exchangeable observations, many practical problems involve multiple related groups that are expected to share statistical structure. The Hierarchical Dirichlet Process (HDP) \cite{teh2006hierarchical} extends the DP to such settings by introducing a hierarchy of random measures. Specifically, a global Dirichlet process generates a shared set of atoms, and group-specific Dirichlet processes are drawn conditionally on this global measure. This hierarchical construction enables different groups to share mixture components while retaining group-level variability. As a result, the HDP provides a natural nonparametric prior for grouped data and has become a foundational tool for multi-level clustering problems.

Most existing applications of the HDP focus on text modeling and topic discovery \cite{teh2006hierarchical}. In particular, the hierarchical Dirichlet process topic model typically relies on the Dirichlet–multinomial conjugate pair. Documents are represented using the bag-of-words paradigm, where word occurrences are modeled by a multinomial likelihood, and the corresponding prior is chosen as a Dirichlet distribution. The conjugacy between the Dirichlet prior and the multinomial likelihood leads to tractable inference and closed-form expressions that facilitate posterior computation.

However, the HDP itself is a highly general Bayesian nonparametric framework and is not restricted to the Dirichlet–multinomial setting. More broadly, analytic tractability can be achieved whenever conjugate prior–likelihood pairs are employed. In particular, the exponential family of distributions provides a unified and elegant class of models encompassing many commonly used distributions, such as the Poisson, Gaussian, Gamma, Bernoulli, and exponential distributions \cite{efron2022exponential}. Conjugate pairs within the exponential family play a central role in Bayesian modeling and arise naturally in numerous applications. For many of these conjugate structures, analytic forms can in principle be derived within the HDP framework.

Although some prior work has acknowledged the broad applicability of exponential-family conjugacy in hierarchical Dirichlet processes \cite{teh2007dirichlet}, explicit analytic expressions are rarely presented in detail. In this paper, we focus on two important members of the exponential family: the Poisson distribution and the normal distribution. We derive explicit analytic results for the corresponding Gamma–Poisson and Normal–Gamma–Normal conjugate pairs within the HDP framework and provide detailed proofs. These results clarify the underlying mathematical structure and offer practical tools for researchers seeking to apply the HDP beyond the traditional Dirichlet–multinomial setting.

\section{Background}

\subsection{Dirichlet Processes}

Let $(\Theta,\mathcal{B})$ be a measurable space and $H$ a probability measure on $\Theta$. 
A Dirichlet Process (DP) with concentration parameter $\gamma>0$ and base measure $H$ is a distribution over random probability measures $G$ such that for any finite measurable partition 
$\{A_1,\dots,A_r\}$ of $\Theta$,
\[
(G(A_1),\dots,G(A_r))
\sim
\text{Dirichlet}(\gamma H(A_1),\dots,\gamma H(A_r)).
\]
We write
\[
G \sim DP(\gamma,H).
\]

A key property of the DP is discreteness: with probability one,
\[
G = \sum_{k=1}^{\infty} \beta_k \delta_{\phi_k},
\]
where $\{\phi_k\}$ are i.i.d.\ draws from $H$ and $\{\beta_k\}$ are random weights summing to one.
This discreteness induces clustering among samples drawn from $G$.

Given $G \sim DP(\gamma,H)$ and parameters
\[
\theta_i \mid G \sim G,
\]
the predictive distribution admits the well-known Pólya urn representation,
\[
\theta_i \mid \theta_{1:i-1}
\sim
\sum_{k=1}^{K}
\frac{n_k}{i-1+\gamma} \delta_{\phi_k}
+
\frac{\gamma}{i-1+\gamma} H,
\]
where $n_k$ is the number of previous observations assigned to atom $\phi_k$ and $K$ is the number of distinct values among $\{\theta_1,\dots,\theta_{i-1}\}$.

\subsection{Hierarchical Dirichlet Processes}

The Hierarchical Dirichlet Process (HDP) extends the DP to grouped data.
Suppose observations are divided into groups indexed by $j=1,\dots,J$.
The HDP introduces a global random measure $G_0$ and group-specific random measures $\{G_j\}$:

\begin{align}
G_0 \mid \gamma, H &\sim DP(\gamma, H), \\
G_j \mid \alpha_0, G_0 &\sim DP(\alpha_0, G_0), \\
\theta_{ji} \mid G_j &\sim G_j, \\
x_{ji} \mid \theta_{ji} &\sim p(x_{ji}\mid \theta_{ji}).
\end{align}

Because $G_0$ is discrete almost surely, the group-level measures $G_j$ share the same atoms $\{\phi_k\}$, but with group-specific weights.
Consequently, mixture components are shared across groups while allowing group-level variability.

\subsection{Chinese Restaurant Franchise}

The Chinese Restaurant Franchise (CRF) provides a constructive representation of the HDP.
Within each group $j$, customers (observations) sit at tables, and each table serves a global dish.

Let $t_{ji}$ denote the table assignment of customer $i$ in group $j$, and let $\psi_{jt}$ denote the dish assigned to table $t$ in group $j$.
The CRF predictive rules are

\begin{align}
\theta_{ji} \mid \text{other } \theta, \alpha_0, G_0 
&\sim 
\sum_{t=1}^{m_{j.}}
\frac{n_{jt.}}{i-1+\alpha_0}
\delta(\psi_{jt})
+
\frac{\alpha_0}{i-1+\alpha_0} G_0, \\
\psi_{jt} \mid \text{other } \psi, \gamma, H 
&\sim
\sum_{k=1}^{K}
\frac{m_{.k}}{m_{..}+\gamma}
\delta(\phi_k)
+
\frac{\gamma}{m_{..}+\gamma} H.
\end{align}

Here,
\[
n_{jt.} = \text{number of customers at table } t \text{ in group } j,
\]
\[
m_{.k} = \text{number of tables serving dish } k,
\qquad
m_{..} = \sum_k m_{.k}.
\]

The first equation describes how a new customer chooses an existing table or starts a new table.
The second equation describes how a new table selects an existing global dish or draws a new dish from the base distribution $H$.
This two-level clustering mechanism induces shared mixture components across groups while maintaining within-group flexibility.

\section{Main Results}

\subsection{Notation}

We consider grouped observations under the hierarchical Dirichlet process (HDP) framework. 
The following notation will be used throughout the paper.

\begin{itemize}

\item \textbf{Observed data.}  
For group $j = 1,\dots,J$ and observation $i = 1,\dots,n_j$,
\[
x_{ji} \in \mathbb{R}
\quad \text{or} \quad
\boldsymbol{x}_{ji} \in \mathbb{R}^d
\]
denotes the $i$-th observation in group $j$.  
We write $x_{-ji}$ to denote all observations except $x_{ji}$.

\item \textbf{Global atoms (dishes).}  
$\phi_k$ denotes the $k$-th global parameter (dish), drawn from the base measure $H$.

\item \textbf{Table-level parameters.}  
$\psi_{jt}$ denotes the parameter (dish) assigned to table $t$ in group $j$.

\item \textbf{Customer-level parameters.}  
$\theta_{ji}$ denotes the parameter associated with observation $x_{ji}$.

\item \textbf{Indices.}
\begin{itemize}
    \item $t_{ji} \in \{1,\dots,T_j\}$: table index at which customer $(j,i)$ sits,
    \item $k_{jt} \in \{1,\dots,K\}$: dish index served at table $(j,t)$,
    \item $z_{ji} \in \{1,\dots,K\}$: dish index assigned to customer $(j,i)$.
\end{itemize}

These quantities satisfy
\[
\theta_{ji} = \psi_{j t_{ji}} = \phi_{z_{ji}}.
\]

\item \textbf{Counts.}
\begin{itemize}
    \item $n_{jtk}$: number of customers in group $j$ sitting at table $t$ and assigned to dish $k$,
    \item $m_{jk}$: number of tables in group $j$ serving dish $k$.
\end{itemize}

\item \textbf{Index sets.}

\begin{itemize}
    \item Customers (excluding $(j,i)$) assigned to dish $k$:
    \[
    V_{k}^{-ji} 
    :=
    \{(j',i') \neq (j,i) : z_{j'i'} = k\}.
    \]

    \item Customers at table $(j,t)$:
    \[
    T_{jt}
    :=
    \{(j',i') : j'=j,\; t_{j'i'} = t\}.
    \]

    \item Customers (excluding those at table $(j,t)$) assigned to dish $k$:
    \[
    W_{k}^{-jt}
    :=
    \{(j',i') : (j',t_{j'i'}) \neq (j,t),\; z_{j'i'} = k\}.
    \]
\end{itemize}
We use the abbreviations $V$, $T$ and $W$ when there is no ambiguity.

\item \textbf{Likelihood and prior.}

The data-generating model (likelihood) is denoted by
\[
p(x \mid \theta),
\]
and $h(\phi)$ denotes the conjugate prior density for $\phi$.

\item \textbf{Customer-level predictive component.}

For an existing dish $k$, define
\begin{align}
g_c(\phi_k)
&=
h(\phi_k)
\prod_{(j',i') \in V_k^{-ji}}
p(x_{j'i'} \mid \phi_k).
\end{align}

The predictive probability of assigning $x_{ji}$ to dish $k$ is
\begin{align}
p_c(x_{ji} \to k)
=
\frac{
\int
p(x_{ji} \mid \phi_k)\,
g_c(\phi_k)\,
d\phi_k
}{
\int
g_c(\phi_k)\,
d\phi_k
}.
\end{align}

\item \textbf{Table-level predictive component.}

For table $(j,t)$ and dish $k$, define
\begin{align}
g_t(\phi_k)
&=
h(\phi_k)
\prod_{(j',i') \in W_k^{-jt}}
p(x_{j'i'} \mid \phi_k).
\end{align}

Let $x_{jt} := \{x_{ji} : t_{ji}=t\}$ denote all observations at table $(j,t)$.  
Then the predictive probability that table $(j,t)$ is assigned to dish $k$ is
\begin{align}
p_t(x_{jt} \to k)
=
\frac{
\int
\left[
\prod_{(j,i)\in T_{jt}}
p(x_{ji} \mid \phi_k)
\right]
g_t(\phi_k)
d\phi_k
}{
\int
g_t(\phi_k)
d\phi_k
}.
\end{align}

\item \textbf{New dish case.}

We denote by $k^*$ a new dish index for which no observations are currently assigned.  
The quantities $p_c(x_{ji} \to k^*)$ and $p_t(x_{jt} \to k^*)$ are defined analogously with
\[
g(\phi_{k^*}) = h(\phi_{k^*}).
\]

\item \textbf{Remark.}

The functions $g_c(\phi_k)$ and $g_t(\phi_k)$ are unnormalized densities.  
After normalization,
\[
\frac{g(\phi_k)}{\int g(\phi_k)\, d\phi_k}
\]
defines the posterior distribution of $\phi_k$ given the corresponding data.  
Accordingly, $p_c(x_{ji} \to k)$ and $p_t(x_{jt} \to k)$ are posterior predictive densities.

\end{itemize}

For clarity of presentation, the main notation used throughout the paper is summarized in Table~\ref{tab:notation}.

\begin{table}[ht]
\centering
\caption{Summary of notation used in the HDP framework}\label{tab:notation}
\begin{tabular}{ll}
\hline
\textbf{Symbol} & \textbf{Description} \\
\hline

$x_{ji}$ or $\boldsymbol{x}_{ji}$ 
& $i$-th observation in group $j$ \\

$x_{-ji}$ 
& All observations except $x_{ji}$ \\

$\phi_k$ 
& $k$-th global parameter (dish), drawn from base measure $H$ \\

$\psi_{jt}$ 
& Parameter (dish) assigned to table $t$ in group $j$ \\

$\theta_{ji}$ 
& Parameter associated with observation $x_{ji}$ \\

$t_{ji}$ 
& Table index of customer $(j,i)$ \\

$z_{ji}$ 
& Dish index assigned to customer $(j,i)$ \\

$k_{jt}$ 
& Dish index served at table $(j,t)$ \\

$n_{jtk}$ 
& Number of customers in group $j$, table $t$, dish $k$ \\

$m_{jk}$ 
& Number of tables in group $j$ serving dish $k$ \\

$V_k^{-ji}$ 
& Customers (excluding $(j,i)$) assigned to dish $k$ \\

$T_{jt}$ 
& Customers seated at table $(j,t)$ \\

$W_k^{-jt}$ 
& Customers (excluding table $(j,t)$) assigned to dish $k$ \\

$p(x \mid \theta)$ 
& Likelihood function \\

$h(\phi)$ 
& Conjugate prior density \\

$g_c(\phi_k)$ 
& Unnormalized posterior for $\phi_k$ (customer-level) \\

$g_t(\phi_k)$ 
& Unnormalized posterior for $\phi_k$ (table-level) \\

$p_c(x_{ji} \to k)$ 
& Customer-level posterior predictive density \\

$p_t(x_{jt} \to k)$ 
& Table-level posterior predictive density \\

$k^*$ 
& Index of a new dish \\

\hline
\end{tabular}
\end{table}

\subsection{Gibbs Sampling and General Representation of the Predictive Components}

In this paper we consider the Chinese Restaurant Franchise representation of HDP, and Gibbs sampling is one of the most popular inference technique. In this subsection we briefly review related results in the original work of \cite{teh2006hierarchical}.

Gibbs sampling provides a tractable Markov chain Monte Carlo (MCMC) procedure for posterior inference in hierarchical Dirichlet processes. Under the Chinese Restaurant Franchise (CRF) representation, the infinite-dimensional random measures are marginalized out, and inference is performed by iteratively sampling the latent table assignments, dish assignments, and global parameters from their full conditional distributions. This collapsed representation exploits conjugacy and leads to simple conditional update rules involving predictive probabilities. These predictive components play a central role in both the computational algorithm and the analytic results derived in the subsequent sections.

It is shown that:
\begin{align}
& p(t_{ji} = t \mid \boldsymbol{t} \setminus t_{ji}, \boldsymbol{k}, \boldsymbol{\phi}, \mathbf{x}) \propto 
\begin{cases} 
\alpha_0 p(x_{ji} \mid \phi_{k_{jt^{*}}}) & \text{if } t = t^{*} \\
n_{jt}^{-i} p(x_{ji} \mid \phi_{k_{jt}}) & \text{if } t \text{ currently used}
\end{cases} \label{eq:p_tji} \\[6pt]
& p(k_{jt} = k \mid \boldsymbol{t}, \boldsymbol{k} \setminus k_{jt}, \boldsymbol{\phi}, \mathbf{x}) \propto 
\begin{cases} 
\gamma \prod_{i:t_{ji}=t} p(x_{ji} \mid \phi_k) & \text{if } k = k^{*} \\
m_k^{-t} \prod_{i:t_{ji}=t} p(x_{ji} \mid \phi_k) & \text{if } k \text{ currently used}
\end{cases} \\[6pt]
& p(\phi_{k} \mid \boldsymbol{t}, \boldsymbol{k}, \boldsymbol{\phi} \setminus \phi_{k}, \mathbf{x}) 
\propto h(\phi_{k}) \prod_{ji:k_{jt_{ji}}=k} p(x_{ji} \mid \phi_{k}) 
\end{align}

When $H$ is conjugate to $p(\cdot)$, we can marginalize out $\boldsymbol{\phi}$. For example, in Eq~\ref{eq:p_tji} when $t$ is currently used, we have:

\begin{align*}
&p(t_{ji} = t \mid \boldsymbol{t} \setminus t_{ji}, \boldsymbol{k},  \mathbf{x})\\
= &
\int p(t_{ji} = t, \boldsymbol{\phi} \mid \boldsymbol{t} \setminus t_{ji}, \boldsymbol{k}, \mathbf{x}) d \boldsymbol{\phi} \\
= &
\int p(t_{ji} = t \mid \boldsymbol{\phi}, \boldsymbol{t} \setminus t_{ji}, \boldsymbol{k}, \mathbf{x}) p( \boldsymbol{\phi} \mid \boldsymbol{t} \setminus t_{ji}, \boldsymbol{k}, \mathbf{x}) d \boldsymbol{\phi} \\
\propto & \int n_{jt}^{-i} p(x_{ji} \mid \phi_{k_{jt}}) h(\phi_{k_{jt}}) \prod_{j'i':j'i'\neq ji,k_{j't_{j'i'}}=k_{jt}} p(x_{ji} \mid \phi_{k_{jt}}) d \phi_{k_{jt}}  \\
\propto & n_{jt}^{-i} \int p(x_{j'i'} \mid \phi_{k_{jt}}) h(\phi_{k_{jt}}) \prod_{j'i':j'i'\neq ji,t_{j'i'}=t} p(x_{j'i'} \mid \phi_{k_{jt}}) d \phi_{k_{jt}}
\end{align*}

The integration part is exactly the definition of $p_t(x_{jt} \rightarrow k)$, i.e., the customer-level predictive component. Similar analysis can be applied to the customer-level mixture component $p_c(x_{ji} \rightarrow k)$.

From the derivation, we can observe the central role of the predictive mixture components in Gibbs sampling for HDP inference. Next we will present analytical results of two types of conjugate pairs for the predictive mixture components.

\subsection{Gamma–Poisson Conjugate Pair}

The Gamma–Poisson conjugate pair is widely used for modeling count data arising in applications such as text analysis, event monitoring, and traffic modeling. 
Under the HDP framework, this structure facilitates flexible clustering of heterogeneous count data across groups.

In this section we define:
\begin{align*}
\alpha_v 
&= \alpha + \sum_{(j',i') \in V_k^{-ji}} x_{j'i'}, 
&
\beta_v 
= \beta + |V_k^{-ji}|, 
\\[6pt]
\alpha_w 
&= \alpha + \sum_{(j',i') \in W_k^{-jt}} x_{j'i'}, 
&
\beta_w 
= \beta + |W_k^{-jt}|.
\end{align*}

The Gamma–Poisson conjugate pair is defined as:

\begin{align*}
p(x|\phi_k) \sim Poisson(\phi_k) \\
h(\phi_k) \sim Gamma(\alpha, \beta)
\end{align*}

\[
p(x \mid \phi_k) = \Pr(X = x \mid \phi_k)
= \frac{\phi_k^{\,x} e^{-\phi_k}}{x!},
\qquad x = 0,1,2,\dots
\]

\[
h(\phi_k)
= \frac{\beta^{\alpha}}{\Gamma(\alpha)} \,
\phi_k^{\alpha - 1} e^{-\beta \phi_k},
\qquad \phi_k > 0
\]

\subsubsection{Customer-Level Predictive Component}

\begin{align*}
g_c(\phi_k) =& h(\phi_k) \prod_{j'i' \neq ji, z_{j'i'} = k} p(x_{j'i'}|\phi_k) = \frac{\beta^{\alpha}}{\Gamma(\alpha)} \,
\phi_k^{\alpha - 1} e^{-\beta \phi_k} \prod_{j'i' \in V} \frac{\phi_k^{\,x_{j'i'}} e^{-\phi_k}}{x_{j'i'}!}
\end{align*}

\begin{align*}
\int g_c(\phi_k) d \phi_k =& \int \frac{\beta^{\alpha}}{\Gamma(\alpha)} \,
\phi_k^{\alpha - 1} e^{-\beta \phi_k} \prod_{j'i' \in V} \frac{\phi_k^{\,x_{j'i'}} e^{-\phi_k}}{x_{j'i'}!} d \phi_k \\
=&  \frac{1}{\prod_{j'i' \in V} x_{j'i'}!} \frac{\beta^{\alpha}}{\Gamma(\alpha)} \,\int 
\phi_k^{\alpha_v - 1} e^{-\beta_v \phi_k} d \phi_k \\
=&  \frac{1}{\prod_{j'i' \in V} x_{j'i'}!} \frac{\beta^{\alpha}}{\Gamma(\alpha)} \,\frac{\Gamma(\alpha_v)}{\beta^{\alpha_v}} \\
\end{align*}

\begin{align*}
& \int p(x_{ji}|\phi_k)g_c(\phi_k)d\phi_k\\
= & \int \frac{\phi_k^{\,x_{ji}} e^{-\phi_k}}{x_{ji}!} \frac{\beta^{\alpha}}{\Gamma(\alpha)} \,
\phi_k^{\alpha - 1} e^{-\beta \phi_k} \prod_{j'i' \in V} \frac{\phi_k^{\,x_{j'i'}} e^{-\phi_k}}{x_{j'i'}!} d\phi_k\\
= & \frac{1}{x_{ji}!} \frac{\beta^{\alpha}}{\Gamma(\alpha)} \frac{1}{\prod_{j'i' \in V}x_{j'i'}!} \, \int \phi_k^{\,x_{ji}} e^{-\phi_k}
\phi_k^{\alpha - 1} e^{-\beta \phi_k} \prod_{j'i' \in V} {\phi_k^{\,x_{j'i'}} e^{-\phi_k}}d\phi_k\\
= & \frac{1}{x_{ji}!} \frac{\beta^{\alpha}}{\Gamma(\alpha)} \frac{1}{\prod_{j'i' \in V}x_{j'i'}!} \, \int \phi_k^{\,x_{ji}+\alpha_v-1} 
 e^{-(\beta_v+1) \phi_k} d\phi_k\\
= & \frac{1}{x_{ji}!} \frac{\beta^{\alpha}}{\Gamma(\alpha)} \frac{1}{\prod_{j'i' \in V}x_{j'i'}!} \, \frac{\Gamma(x_{ji}+\alpha_v)}{(\beta_v+1)^{x_{ji}+\alpha_v}}
\end{align*}

Finally, we have:
\begin{align*}
p_c(x_{ji} \to k)
&=
\frac{
\int
p(x_{ji} \mid \phi_k)\,
g_c(\phi_k)\,
d\phi_k
}{
\int
g_c(\phi_k)\,
d\phi_k
}\\
&=\frac{1}{x_{ji}!}
\frac{\Gamma(x_{ji}+\alpha_v)}{\Gamma(\alpha_v)}
\frac{\beta_v^{\alpha_v}}
{(\beta_v+1)^{x_{ji}+\alpha_v}}.
\end{align*}

\subsubsection{Table-Level Predictive Component}

\begin{align*}
\int g_t(\phi_k) d \phi_k =& \int \frac{\beta^{\alpha}}{\Gamma(\alpha)} \,
\phi_k^{\alpha - 1} e^{-\beta \phi_k} \prod_{j'i' \in W} \frac{\phi_k^{\,x_{j'i'}} e^{-\phi_k}}{x_{j'i'}!} d \phi_k \\
=&  \frac{1}{\prod_{j'i' \in W} x_{j'i'}!} \frac{\beta^{\alpha}}{\Gamma(\alpha)} \,\int 
\phi_k^{\alpha_w - 1} e^{-\beta_w \phi_k} d \phi_k \\
=&  \frac{1}{\prod_{j'i' \in V} x_{j'i'}!} \frac{\beta^{\alpha}}{\Gamma(\alpha)} \,\frac{\Gamma(\alpha_w)}{\beta^{\alpha_w}} 
\end{align*}

\begin{align*}
& \int p(x_{jt}|\phi_k)g_t(\phi_k)d\phi_k
= \int \prod_{ji \in T} p(x_{ji}|\phi_k)g_t(\phi_k)d\phi_k\\
= & \int \prod_{ji \in T}  \frac{\phi_k^{\,x_{ji}} e^{-\phi_k}}{x_{ji}!} \left [ \frac{\beta^{\alpha}}{\Gamma(\alpha)} \,
\phi_k^{\alpha - 1} e^{-\beta \phi_k} \prod_{j'i' \in W} \frac{\phi_k^{\,x_{j'i'}} e^{-\phi_k}}{x_{j'i'}!} d \right ] \phi_k\\
= & \frac{1}{\prod_{ji \in T} x_{ji}!} \frac{\beta^{\alpha}}{\Gamma(\alpha)} \frac{1}{\prod_{j'i' \in W}x_{j'i'}!} \, \int \phi_k^{\,\sum_T x_{ji}} e^{-|T|\phi_k}
\phi_k^{\alpha - 1} e^{-\beta \phi_k} \prod_{j'i' \in V} {\phi_k^{\,x_{j'i'}} e^{-\phi_k}}d\phi_k\\
= & \frac{1}{\prod_{ji \in T} x_{ji}!} \frac{\beta^{\alpha}}{\Gamma(\alpha)} \frac{1}{\prod_{j'i' \in V}x_{j'i'}!} \, \int \phi_k^{\,\sum_T x_{ji}+\alpha_w-1} 
 e^{-(\beta_w+|T|) \phi_k} d\phi_k\\
= & \frac{1}{x_{ji}!} \frac{\beta^{\alpha}}{\Gamma(\alpha)} \frac{1}{\prod_{j'i' \in V}x_{j'i'}!} \, \frac{\Gamma(\sum_T x_{ji}+\alpha_v)}{(\beta_v+|T|)^{\sum_T x_{ji}+\alpha_v}}
\end{align*}

Finally, we have:
\begin{align*}
p_t(x_{jt} \rightarrow k) & = \frac{\int g_t(\phi_k) p(x_{jt}|\phi_k)d\phi_k}{\int g_t(\phi_k)d\phi_k} = \frac{\int g_t(\phi_k) \prod_{ji:t_{ji}=t} p(x_{ji}|\phi_k)d\phi_k}{\int g_t(\phi_k)d\phi_k}\\
&=
\frac{\Gamma(\sum_{T} x_{ji}+\alpha_w)}
{\Gamma(\alpha_w)\prod_{T} x_{ji}!}
\frac{\beta_w^{\alpha_w}}
{(\beta_w+|T|)^{\sum_{T} x_{ji}+\alpha_w}}.
\end{align*}

\subsection{Normal–Gamma–Normal Conjugate Pair}

The Normal–Gamma–Normal conjugate structure provides a natural Bayesian model for continuous multivariate data with unknown mean and precision. 
Under the HDP framework, this conjugate pair enables nonparametric clustering of grouped Gaussian observations while preserving closed-form posterior and predictive distributions.

In this section we define the posterior hyperparameters based on the sufficient statistics of the observations assigned to dish $k$. Specifically, we define:
\begin{align*}
V_k^{-ji} &= \{(j',i') \neq (j,i): z_{j'i'} = k\}, & n_v &= |V_k^{-ji}|, \\
\bar x_v &= \frac{1}{n_v} \sum_{(j',i') \in V_k^{-ji}} x_{j'i'}, & 
S_v &= \sum_{(j',i') \in V_k^{-ji}} \| x_{j'i'} - \bar x_v \|^2, \\[2mm]
W_k^{-jt} &= \{(j',i'): z_{j'i'} = k, t_{j'i'} \neq t\}, & n_w &= |W_k^{-jt}|, \\
\bar x_w &= \frac{1}{n_w} \sum_{(j',i') \in W_k^{-jt}} x_{j'i'}, &
S_w &= \sum_{(j',i') \in W_k^{-jt}} \| x_{j'i'} - \bar x_w \|^2.
\end{align*}

The conjugate pair is given by
\begin{align*}
p(x \mid \mu_k,\lambda_k) 
&\sim \mathcal N(\mu_k, \lambda_k^{-1}\mathbf I), \\
h(\mu_k,\lambda_k) 
&\sim NG(\mu_0,\kappa_0,\alpha_0,\beta_0).
\end{align*}

\[
p(x \mid \mu_k,\lambda_k)
=
\left(\frac{\lambda_k}{2\pi}\right)^{\frac d2}
\exp\!\left(
-\frac{\lambda_k}{2}\|x-\mu_k\|^2
\right).
\]

\[
h(\mu_k,\lambda_k)
=
\left(\frac{\kappa_0\lambda_k}{2\pi}\right)^{\frac d2}
\frac{\beta_0^{\alpha_0}}{\Gamma(\alpha_0)}
\lambda_k^{\alpha_0-1}
\exp\!\left(
-\frac{\kappa_0\lambda_k}{2}\|\mu_k-\mu_0\|^2
-\beta_0\lambda_k
\right).
\]

\subsubsection{Conditional Posterior of the Precision Parameter}

First, we derive the conditional posterior of the precision parameter $\lambda_k$, in both the customer-level and the table-level formulations.

\textbf{Customer-Level Posterior.}
The posterior kernel excluding observation $(j,i)$ is
\[
g_c(\mu_k,\lambda_k)
\propto
\prod_{(j',i')\in V_k^{-ji}}
p(x_{j'i'}\mid\mu_k,\lambda_k)
\,h(\mu_k,\lambda_k).
\]

Expanding the likelihood product,
\[
\prod_{V_k^{-ji}}
p(x_{j'i'}\mid\mu_k,\lambda_k)
=
\left(\frac{\lambda_k}{2\pi}\right)^{\frac{d}{2}n_v}
\exp\!\left(
-\frac{\lambda_k}{2}
\sum_{V_k^{-ji}}
\|x_{j'i'}-\mu_k\|^2
\right).
\]

Thus,
\[
g_c(\mu_k,\lambda_k)
\propto
\lambda_k^{\alpha_0-1+\frac d2 n_v}
\exp\!\left(
-\frac{\lambda_k}{2}
\sum_{V_k^{-ji}}\|x_{j'i'}-\mu_k\|^2
-\frac{\kappa_0\lambda_k}{2}\|\mu_k-\mu_0\|^2
-\beta_0\lambda_k
\right).
\]

Using the identity
\[
\sum_{V_k^{-ji}}\|x-\mu\|^2
=
S_v
+
n_v\|\mu-\bar x_v\|^2,
\]
and completing the square,
\[
n_v\|\mu_k-\bar x_v\|^2
+
\kappa_0\|\mu_k-\mu_0\|^2
=
\kappa_v\|\mu_k-\mu_v\|^2
+
\frac{\kappa_0 n_v}{\kappa_v}
\|\bar x_v-\mu_0\|^2,
\]
where
\[
\kappa_v=\kappa_0+n_v,
\qquad
\mu_v=\frac{\kappa_0\mu_0+n_v\bar x_v}{\kappa_v}.
\]

Integrating over $\mu_k$,
\[
\int
\exp\!\left(
-\frac{\kappa_v\lambda_k}{2}
\|\mu_k-\mu_v\|^2
\right)
d\mu_k
=
\left(\frac{2\pi}{\kappa_v\lambda_k}\right)^{\frac d2}.
\]

Hence,
\[
g_c(\lambda_k)
\propto
\lambda_k^{\alpha_v-1}
\exp(-\beta_v\lambda_k),
\]
with
\[
\alpha_v=\alpha_0+\frac d2 n_v,
\]
\[
\beta_v
=
\beta_0
+\frac12 S_v
+\frac{\kappa_0 n_v}{2\kappa_v}
\|\bar x_v-\mu_0\|^2.
\]

Therefore,
\[
\lambda_k \mid V_k^{-ji}
\sim
Gamma(\alpha_v,\beta_v).
\]

\textbf{Table-Level Posterior.}
The posterior is
\[
g_t(\mu_k,\lambda_k)
\propto
\prod_{W_k^{-jt}}
p(x_{j'i'}\mid\mu_k,\lambda_k)
\,h(\mu_k,\lambda_k).
\]

Repeating the same derivation yields
\[
\kappa_w=\kappa_0+n_w,
\qquad
\mu_w=\frac{\kappa_0\mu_0+n_w\bar x_w}{\kappa_w},
\]
\[
\alpha_w=\alpha_0+\frac d2 n_w,
\]
\[
\beta_w
=
\beta_0
+\frac12 S_w
+\frac{\kappa_0 n_w}{2\kappa_w}
\|\bar x_w-\mu_0\|^2.
\]

Hence,
\[
\lambda_k \mid W_k^{-jt}
\sim
Gamma(\alpha_w,\beta_w).
\]

\subsubsection{Customer-Level Predictive Component}

The customer-level predictive component is obtained by integrating the likelihood of
$x_{ji}$ with respect to the posterior kernel $g_c(\mu_k,\lambda_k)$:
\[
p_c(x_{ji} \to k)
=
\int
p(x_{ji}\mid \mu_k,\lambda_k)
g_c(\mu_k,\lambda_k)
\, d\mu_k d\lambda_k .
\]

Using the posterior form derived above,
\[
\mu_k \mid \lambda_k, V_k^{-ji}
\sim
\mathcal N\!\left(\mu_v, (\kappa_v\lambda_k)^{-1}\mathbf I \right),
\qquad
\lambda_k \mid V_k^{-ji}
\sim
Gamma(\alpha_v,\beta_v).
\]

First integrate over $\mu_k$. Since both densities are Gaussian in $\mu_k$,
\[
\int
p(x_{ji}\mid\mu_k,\lambda_k)
p(\mu_k\mid\lambda_k,V_k^{-ji})
\, d\mu_k
=
\left(
\frac{\lambda_k}{2\pi}
\right)^{\frac d2}
\left(
\frac{\kappa_v}{\kappa_v+1}
\right)^{\frac d2}
\exp\!\left(
-\frac{\lambda_k}{2}
\frac{\kappa_v}{\kappa_v+1}
\|x_{ji}-\mu_v\|^2
\right).
\]

Thus,
\[
p_c(x_{ji} \to k)
=
\int
\left(
\frac{\lambda_k}{2\pi}
\right)^{\frac d2}
\left(
\frac{\kappa_v}{\kappa_v+1}
\right)^{\frac d2}
\exp\!\left(
-\frac{\lambda_k}{2}
\frac{\kappa_v}{\kappa_v+1}
\|x_{ji}-\mu_v\|^2
\right)
\frac{\beta_v^{\alpha_v}}{\Gamma(\alpha_v)}
\lambda_k^{\alpha_v-1}
e^{-\beta_v\lambda_k}
\, d\lambda_k.
\]

Collecting powers of $\lambda_k$,
\[
p_c(x_{ji} \to k)
=
C
\int
\lambda_k^{\alpha_v+\frac d2 -1}
\exp\!\left(
-
\lambda_k
\left[
\beta_v
+
\frac12
\frac{\kappa_v}{\kappa_v+1}
\|x_{ji}-\mu_v\|^2
\right]
\right)
d\lambda_k,
\]
where $C$ gathers constants independent of $\lambda_k$.

Using the Gamma integral identity,
\[
\int_0^\infty
\lambda^{a-1}
e^{-b\lambda}
\, d\lambda
=
\frac{\Gamma(a)}{b^a},
\]
we obtain
\[
p_c(x_{ji} \to k)
=
\frac{\Gamma\!\left(\alpha_v+\frac d2\right)}
{\Gamma(\alpha_v)
(\pi \alpha_v)^{\frac d2}}
\left(
\frac{\kappa_v}{\kappa_v+1}
\right)^{\frac d2}
\left[
1
+
\frac{\kappa_v}{\alpha_v(\kappa_v+1)}
\|x_{ji}-\mu_v\|^2
\right]^{-\left(\alpha_v+\frac d2\right)}.
\]

Therefore,
\[
p_c(x_{ji} \to k)
=
t_{2\alpha_v}
\left(
x_{ji}
\mid
\mu_v,
\frac{\beta_v(\kappa_v+1)}
{\alpha_v\kappa_v}\mathbf I
\right),
\]
which is a multivariate Student-$t$ distribution with $2\alpha_v$ degrees of freedom.

\subsubsection{Table-Level Predictive Component}

The table-level predictive component is obtained analogously:
\[
p_t(x_{jt} \to k)
=
\int
p(x_{jt}\mid \mu_k,\lambda_k)
g_t(\mu_k,\lambda_k)
\, d\mu_k d\lambda_k .
\]

Using the posterior parameters
\[
\kappa_w=\kappa_0+n_w,
\qquad
\mu_w=\frac{\kappa_0\mu_0+n_w\bar x_w}{\kappa_w},
\]
\[
\alpha_w=\alpha_0+\frac d2 n_w,
\qquad
\beta_w
=
\beta_0
+\frac12 S_w
+\frac{\kappa_0 n_w}{2\kappa_w}
\|\bar x_w-\mu_0\|^2,
\]
we integrate in the same manner as above.

The resulting predictive distribution is
\[
p_t(x_{jt} \to k)
=
t_{2\alpha_w}
\left(
x_{jt}
\mid
\mu_w,
\frac{\beta_w(\kappa_w+1)}
{\alpha_w\kappa_w}\mathbf I
\right).
\]

Hence, both customer-level and table-level predictive components admit closed-form Student-$t$ distributions under the Normal–Gamma conjugate structure.

\section{Conclusion}

We presented a unified derivation of collapsed inference for Hierarchical Dirichlet Process mixture models under conjugate exponential-family priors. By explicitly computing the customer-level and table-level posterior and predictive distributions, we showed that closed-form updates arise naturally from sufficient statistics: the Gamma–Poisson model yields Negative Binomial predictives, while the Normal–Gamma–Normal model leads to multivariate Student-$t$ predictives. 
We hope that our rigorous derivations and solid analytical results will benefit future research on HDP models, and further support the application of nonparametric Bayesian clustering methods in increasingly diverse and complex domains.

\bibliographystyle{IEEEtran}
\bibliography{references} 

\end{document}